\documentclass[runningheads]{llncs}
\usepackage{graphicx}
\usepackage{hyperref}
\begin{document}
\title{Evolved Art with Transparent, Overlapping, and Geometric Shapes}
%
%
\author{Joachim Berg \and
Nils Gustav Andreas Berggren \and
Sivert Allergodt Borgeteien \and
Christian Ruben Alexander Jahren \and
Arqam Sajid \and
Stefano Nichele}
\authorrunning{Berg, Berggren, Borgeteien, Jahren, Sajid, and Nichele (2019)}
%
\institute{Oslo Metropolitan University, Faculty of Technology, Art, and Design - AI Lab
\email{stefano.nichele@oslomet.no}}
\maketitle              
\begin{abstract}
In this work, an evolutionary art project is presented where images are approximated by transparent, overlapping and geometric shapes of different types, e.g., polygons, circles, lines. Genotypes representing features and order of the geometric shapes are evolved with a fitness function that has the corresponding pixels of an input image as a target goal. A genotype-to-phenotype mapping is therefore applied to render images, as the chosen genetic representation is indirect, i.e., genotypes do not include pixels but a combination of shapes with their properties. Different combinations of shapes, quantity of shapes, mutation types and populations are tested. The goal of the work herein is twofold: (1) to approximate images as precisely as possible with evolved indirect encodings, (2) to produce visually appealing results and novel artistic styles. 

\end{abstract}
\section{Introduction}

In nature, all species of all organisms evolve through natural selection by passing their traits to the next generation, but with small variations, to increase the next generation's ability to survive, compete, and reproduce \cite{dorin2014biological}. This is called evolution by natural selection and can be seen as an algorithm to search for an increasingly better or fitter solution. The algorithm adapts to the environment by making small changes, or mutations, to its previous solution, and by repeating this process the algorithm will iteratively find an equal or a better solution. The DNA sequences are passed from a parent to a child, and this compressed representation of DNA sequences is referred to as genotype. A genotype is a complete heritable genetic identity used to pass genetic information from one generation to the next generation. It can be seen as the recipe for the phenotype which is the actual visual representation of the organism. By applying this to evolution, one can say that the organism passes on its genotype with small variations to the next generation. While it is the genotype that is passed through the generations, it is the phenotype that is evaluated and subjected to fitness.

\begin{figure}[ht]
    \centering
    \includegraphics[width=0.35\textwidth]{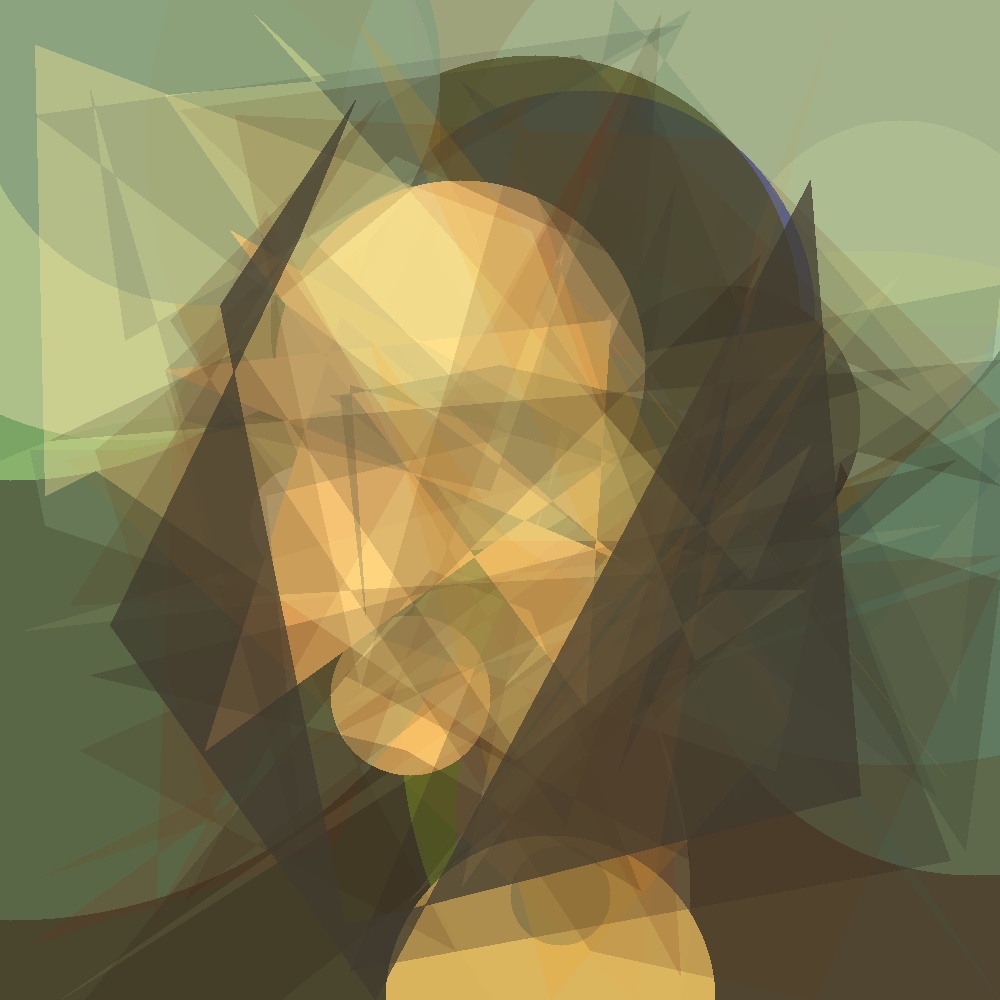}
    \caption{An example of a rendered image genome evolved with the described algorithm.}
    \label{figure:workflow}
\end{figure}

The phenotype's ability to compete, survive, and reproduce is determined by its compatibility with the surrounding environment, i.e., a fitness function. If the child of this phenotype has traits that are less compatible with the environment than its parent, the child's ability to compete and survive may not be enough for it to reproduce. The initial phenotype has to produce children with an equal or better genotype to make sure the newer generations can continue to reproduce.

Using this concept of evolutionary algorithm, we develop a program which uses the same logic as in natural selection, to recreate a target image using arbitrary geometrical shapes such as transparent overlapping circles, polygons, and lines. The chosen geometrical shapes can vary in size, colour, transparency, and placement. In each iteration, or generation, of this process, a collection of shapes is created, and their structures are represented as instance objects with changeable parameters. These objects represent the genes in the genotype. We will refer to this collection of genes as an image genome.

To distinguish a bad solution from a fit solution, a fitness function is needed. If the new solution has a better fitness score than the previous one, the new solution will replace the old one and be used for the next generation. Using this technique the program will find an increasingly better solution, and when the fitness score is good enough or a termination criteria is met, the program will stop. An example of rendered image genome is shown in Figure \ref{figure:workflow}.

Evolutionary art is an active area of research. One of the seminal work in this field is \cite{sims1991artificial} by Karl Sims, where virtual creatures were evolved for the first time. Our work is inspired by \cite{tarimo}, where they evolved images with overlapping polygons with genotypes of fixed size, and the resulting images produce fairly sharp edges. In \cite{barque} an evolutionary algorithm to reconstruct 3D objects based on images is presented. The work in \cite{bergen2009evolving} introduces an user-interactive image evolution system that does not rely on user feedback or supervision. Recent work includes \cite{sipper2019omnirep} for the coevolution of encodings and representations, \cite{paauw} which uses stochastic hillclimbing, simulated annealing and the plant propagation algorithm, \cite{alsing2008genetic} which uses genetic programming, \cite{ashlock2015evolving} for evolving fractal art, and \cite{ashlock2009evolved} for cellular automata-based evolutionary art. Previous work by the authors on evolution of cellular automata structures development and replication include \cite{nichele2018neat}, \cite{nichele2014evolutionary}, \cite{nichele2015morphogenesis}, \cite{nichele2016genotype}.

\section{Methodology}

\subsection{Image genome structure}

\begin{figure}[b]
    \centering
    \includegraphics[width=0.30\textwidth]{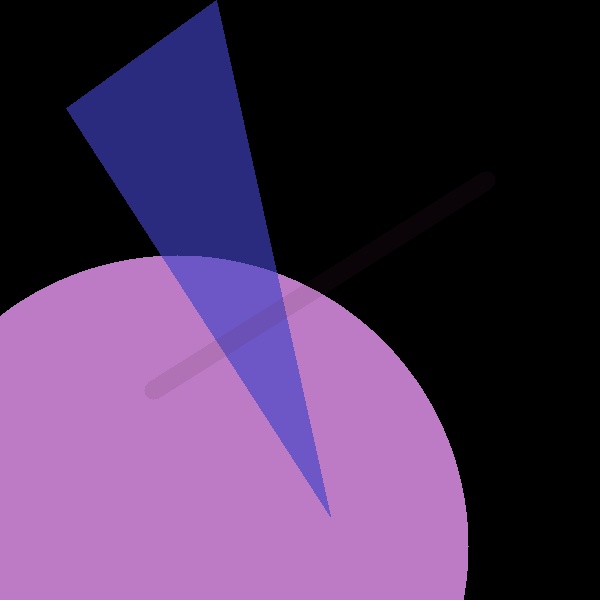}
    \caption{Visual representation of the genes in Table \ref{table:geneval}}
    \label{figure:geneval}
\end{figure}

All image genomes are implemented in the program as objects consisting of genes for each shape. Each gene is made up by parameters that define the size of the canvas, the shape's colour, the transparency/alpha of that shape, and its coordinates on the canvas, and depending on the shape, number of vertices, the radius length and the thickness. A shape's initial gene structure is made up by these parameters: The width and height of the target image, an array for the colours with values from 0-255, a transparency/alpha value between 0-1, and the coordinates as x (value from 0 to width of image) and y (value from 0 to height of image). Additional parameters such as the number of vertices, the length of the radius, and the thickness of the line are added to the end of the gene when it is generated. See Table \ref{table:geneval} for examples for each of the shapes. Their phenotypes are represented visually in Figure \ref{figure:geneval} (left).

\begin{table}[t]
    \caption{Sample gene values}
    \flushleft
    \begin{tabular}{c c c c c c}
        \hline\hline
        Type & Width x Height & Colour & Alpha & Coordinates & Vertices/\\& & & & & Radius/\\& & & & & Thickness\\   [1ex]
        \hline 
        Polygon & 200 x 200 & (65, 6, 197) & 0.64 & [[22, 36], [110, 172], [72, 0]] & 3 \\ [1ex]
        Circle & 200 x 200 & (243, 159, 253) & 0.77 & (59, 182) & 97 \\ [1ex]
        Line & 200 x 200 & (35, 89, 71) & 0.12 & (51, 130), (162, 60) & 6 \\ [1ex] 
    \end{tabular}
    \label{table:geneval}
\end{table}

\subsection{Mutation operations}

Any gene in the genome can be mutated based on a set of given parameters in the program. A modification of a specific or random parameter of the gene is done by retrieving the values of the relevant parameters and modifying them within the limitations given. In our implementation we have included three mutation operations, soft mutation, medium mutation, and a hybrid mutation. The soft mutation updates parameters within a limit, whilst the medium operation replaces existing parameters with new values. The hybrid mutation combines the two former mutations by first doing two soft mutation and then one medium mutation (ratio 2:1). We have also added two mutation factors, probability mutation and chunk mutation. The probability mutation is based on the parameter mutation\_probability, set in the program with a value from 0 to 1. This gives all genes in a genome a probability of mutating decided by the parameter's value. When using the chunk mutation, a number of genes in the genome are always mutated. The number is based on the same parameter, mutation\_probability, but is multiplied with the number of genes in the genome. Example with the probability mutation: If the parameter is set to 0.5, all genes in the genome will have a 50\% probability of mutation. Example with the chunk mutation: If a genome has 100 genes and the variable is set to 0.5, then 50 mutations will take place. This means that there is a probability for the same gene being mutated twice (or more).\footnote{The minimum number of genes to be mutated is 1.}

\subsection{From genotype to phenotype}
To render the genotype's phenotype, a black canvas is created, and then each gene in the genotype is rendered onto the canvas one by one. In our implementation we have used a Python library called OpenCV, as it is capable of rendering shapes with correct alpha values. It also provides simple utility for working with red, green, and blue channels of a 24-bit image. As every gene represents one shape, the circles, polygons, and lines are drawn, filled, and rendered in their respective order, in compliance with the genome structure. The resulting phenotype is evaluated by the fitness function.

\subsection{Fitness function}
The goal of the fitness function is to measure how close the generated image is to the target image, and to distinguish bad solutions from good ones. Our implementation of the fitness function is done by summarising the pixel by pixel difference of the images and that way determine a score. This score is used to determine whether to replace the parent image genome with the child image genome, or discard it. If the score of the child is lower than the parent image genome, the parent is replaced. 

To improve the readability of the fitness score, we convert the absolute score to a relative fitness in percent. The percentage score is calculated by dividing the actual fitness score by the theoretical worst fitness score: The maximum difference (255) in each channel of the image (r, g and b) multiplied by the dimensions of the image in pixels (255 x 3 x width x height). Furthermore, the ratio is converted to a percent and then flipped (100.0\% minus the calculated percentage) to reflect approximation towards 100\% instead of 0\%. The difference decreases as the approximation improves. This way the fitness score is comparable between images of different dimensions and easier to put into context.

The fitness score does not necessarily determine the best rendered image for human eyes. (e.g image with worse score can be more recognisable than another image with better score.) Certain defined features of an image can be more important for recognition by humans.

The program and data from the experiments are available on GitHub\footnote{\url{https://github.com/joacber/Evolved-art-with-transparent-overlapping-and-geometric-shapes}}.

\begin{figure}[t]
    \centering
    \includegraphics[width=0.6\textwidth]{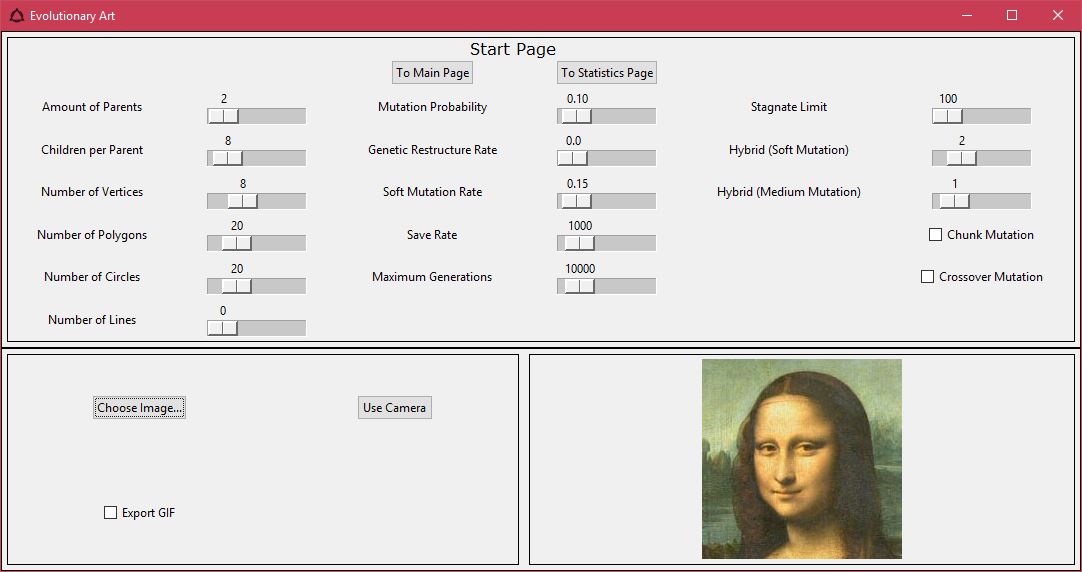}
    \caption{Example of Graphical User Interface (GUI)}
    \label{figure:guifig}
\end{figure}

A graphic interface is available to facilitate the execution of the program, as shown in Figure \ref{figure:guifig}.

\section{Experiments and Results}

\begin{figure}[t]
    \centering
    \includegraphics[width=0.3\textwidth]{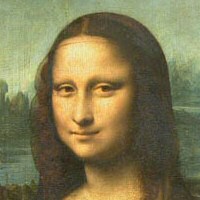} 
    \caption {Target image used in experiments: Mona Lisa by Leonardo.}
    \label{figure:monalisacrop}
\end{figure}

To make the tests comparable, the same target image was used in all experiments. This image was also chosen because it is easily recognisable, see Figure \ref{figure:monalisacrop}. 
The initial parameters for the experiments are summarised in Table \ref{table:inittest}. In particular. the \textit{Number of Parents} parameter reflects how many parents the algorithm has to work with. The best parent has its phenotype shown in the GUI, but every parent's phenotype is saved in the output folder. The \textit{Children per Parent} parameter is the number of children each parent will produce. The minimum value of both is 1, so the program has a population to evolve and compare, and the maximum is set to 100 to limit the processing and the total time it takes. As previously explained, the genome of a genotype consists of a number of genes. The next parameter you see here is \textit{Genes Total}, which is made up of the values of the three parameters below it. The initial number in these tests is 20 genes per genotype. As the experimenting progresses, the number of \textit{Polygons}, \textit{Circles}, and \textit{Lines} will change. The number of \textit{Vertices} for the polygons is an editable parameter, but the radius of the circle and the thickness of the lines are under evolutionary control and are not editable by the user (and therefore not present in the table).

The \textit{Mutation Probability} parameter defines the rate of probability and chunk mutation factors. The \textit{Genetic Restructure Rate} parameter is based on the previous parameter, \textit{Mutation Probability}. If the current number of generations during the process is below one tenth of the maximum number of generations, each gene in the genotype will have 0.1 probability to mutate. The \textit{Soft Mutation Rate} parameter decides the level of change that any gene's parameter can be subjected to. \textit{Hybrid (Medium Mutation)} determines for how many generations in a row the algorithm should run with medium mutation. If it is 0 it only runs soft. \textit{Hybrid (Soft Mutation)} determines how many generations in a row the algorithm should run with soft mutation. If it is 0 and \textit{Hybrid (Medium Mutation)} is not 0 it only runs medium mutation. 

The \textit{Save Rate} parameter defines how often the image is saved (e.g. every 1,000th generation). The \textit{Maximum Generations} is set to 10,000. That means reaching generation number 10,000 is a termination condition, and thus the algorithm will terminate. \textit{Chunk Mutation} decides whether to use chunk mutation as a factor in the mutation operations, instead of probability, which is the standard. The parameter is true or false. The functions of these two mutation factors are explained later. \textit{Crossover Mutation} determines if 2 parents should cross-mutate when producing a child. The parameter is true or false. The child is made up of the main parent's coordinates and shape relevant parameters, and the second parent's colour and alpha values.

\begin{table}[t]
\centering
\caption{Initial Test Parameter Values} 
\begin{tabular}{l l }
\hline\hline
Parameter Name &  Parameter Value \\ [1ex]
\hline
Number of Parents & 1\\ [1ex]
Children per Parent & 1\\ [1ex] 
Genes Total & 20 \\ [1ex]
Polygons & 20 \\ [1ex]
Circles & 0 \\ [1ex] 
Lines & 0 \\ [1ex]
Vertices & 3\\ [1ex]
Mutation Probability & 0.1 \\ [1ex] 
Genetic Restructure Rate & 0 \\ [1ex] 
Soft Mutation Rate & 0.1\\ [1ex]  
Hybrid (Soft Mutation) & 0 \\ [1ex] 
Hybrid (Medium Mutation) & 0 \\ [1ex] 
Chunk Mutation & False\\ [1ex]  
Crossover Mutation & False \\ [1ex]
Save Rate & 1,000 \\ [1ex] 
Maximum Generations & 10,000\\ [1ex] 
\end{tabular}
\label{table:inittest}
\end{table}

\subsection{Number of vertices}

\begin{figure}[t]
    \centering
    \includegraphics[width=\textwidth]{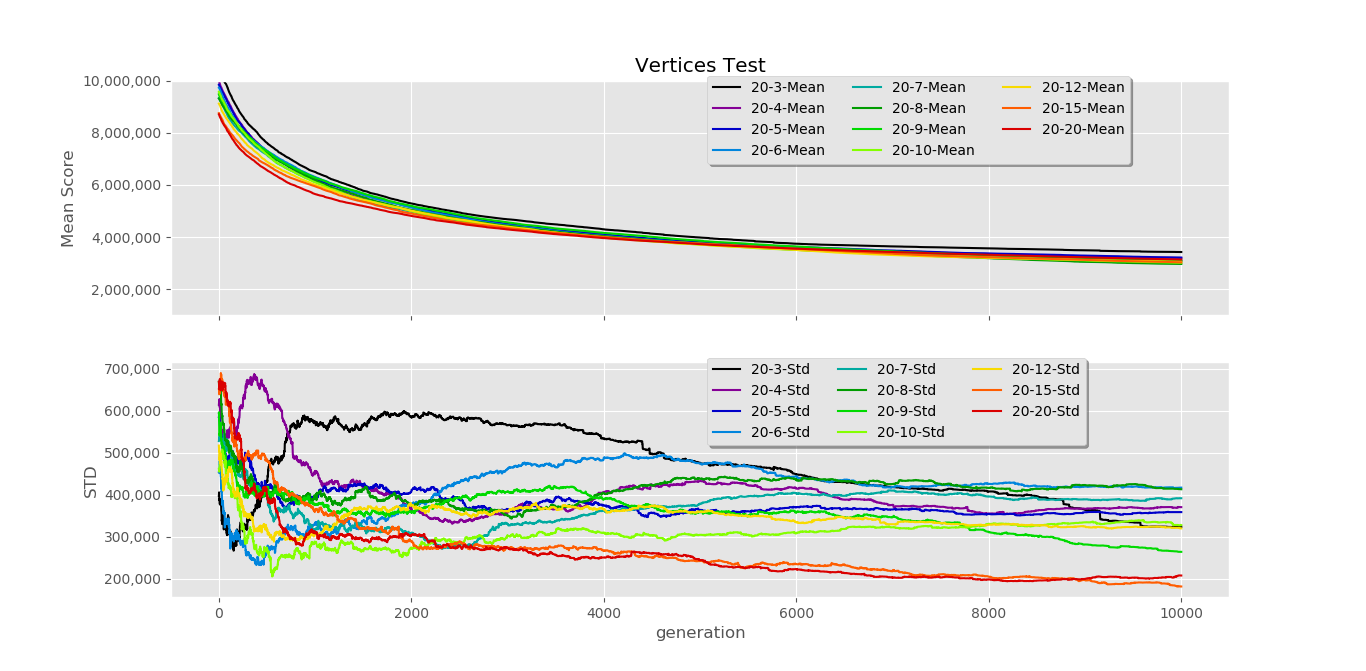} 
    \caption{Results from the vertices test with 20 polygons and a number of vertices. Top graph: average result from all the vertices tests. Bottom graph: standard deviation from the same tests.}
    \label{figure:verticeresultalltests}
\end{figure}

The initial test was to determine the most appropriate number of vertices to use in further testing of polygons. All tests were executed 15 times. Using a genome consisting of 20 polygons, we were able to determine the number of vertices with the most positive impact on the fitness score over 10,000 generations. The best average score was achieved by genomes consisting of polygons with 8 vertices, followed by 10 and 15 (see Table \ref{table:verticesresulttop3}). The top graph in Figure \ref{figure:verticeresultalltests} displays the average results of 15 tests with the number of vertices increasing from 3 to 20. The differences are apparently small. The bottom graph in the same figure displays the standard deviation (STD). If the fitness score has a high variation during the process, the STD value will also be high. Here we can see that the STD value is relatively dynamic in the first 5,000 generations, but flattens out later. This makes sense because in the early stages the image is being created from scratch, and later on the image is mostly being fine-tuned. 

\begin{table}[t]
\caption{Top results from the vertices tests}
\centering
\begin{tabular}{l l l l}
\hline\hline
Rank & Number of Vertices &  Average Score & Average Relative Score\\ [1ex] 
\hline
1 & 8 & 2,885,838 & 90.57\% \\ [1ex]
2 & 10 & 3,014,961 & 90.15\%  \\ [1ex]
3 & 15 & 3,032,196 & 90.09\%  \\ [1ex]
\end{tabular}
\label{table:verticesresulttop3}
\end{table}

\begin{figure}[t]
\centering
    {\includegraphics[width=0.25\textwidth]{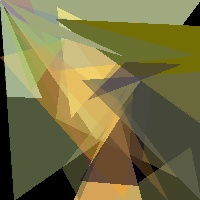}}
    {\includegraphics[width=0.25\textwidth]{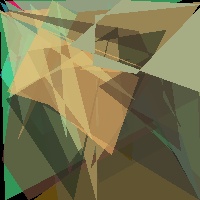} }
    {\includegraphics[width=0.25\textwidth]{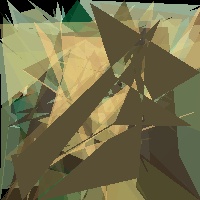} }
    \caption{Samples from the vertices tests (10,000 generations), with 3, 8, and 20 vertices.}
    \label{figure:verticespics}
\end{figure}

\subsection{Polygons}
\begin{figure}[t]
    \centering
    \includegraphics[width=\textwidth]{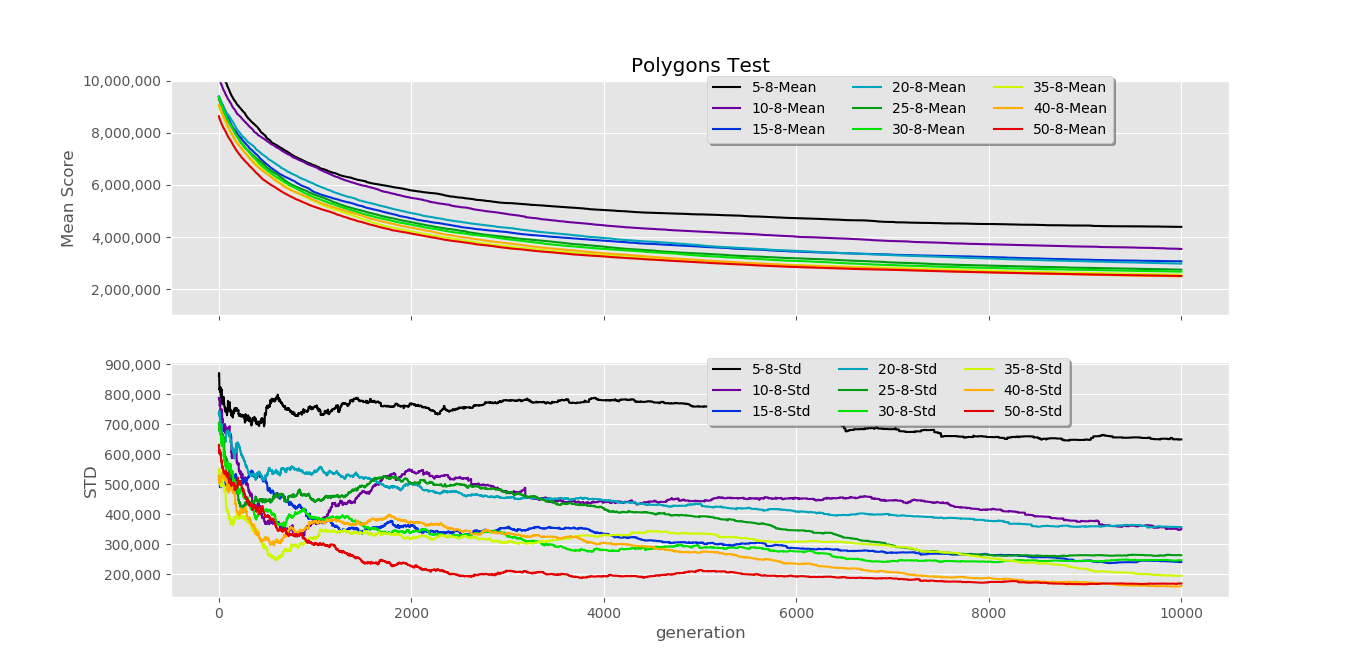} 
    \caption[Results from polygons tests]{Results from polygons tests with 8 vertices and increasing number of polygons. The legend shows number of polygons - number of vertices}
    \label{figure:polytest}
\end{figure}

Using genomes consisting only of polygons with 8 vertices, we found that the fitness score increases almost linearly with an increase in the genome complexity (more polygons). The genome size was increased by 5 after 15 consecutive runs with the same parameters. We saw that the bigger leaps in fitness happened between 5 and 15 polygons (see top graph in Figure \ref{figure:polytest}) - hitting 3,072,442 (89.96\%) average score at 15 contrary to 4,391,165 (85.65\%) at 5. 20 polygons is the milestone hitting 2,996,858 (90.21\%) average score, and 25 polygons subsequently hitting 2,751,771 (91.01\%) average score with an increase of 0.80 percentage points. Increases were minimal with more complex genomes and disproportionate increases in process time up to 40 polygons, with fitness score even dropping between 40 (Avg. Score: 2,494,330) and 50 (Avg. Score: 2,507,859) polygons, thus making 25 polygons the apparent winner for 10,000 generations with an approximation of above 90\% and a relatively big increase (0.80 percent points) from 20 polygons. In the bottom graph of Figure \ref{figure:polytest}, we can see that the STD value flattens out after 4,000 generations.

\begin{figure}[t]
\centering
    {\includegraphics[width=0.25\textwidth]{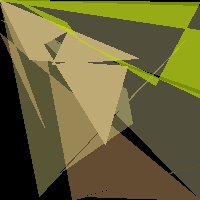}}
    {\includegraphics[width=0.25\textwidth]{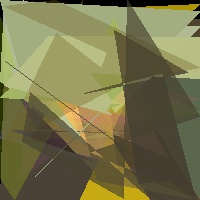} }
    {\includegraphics[width=0.25\textwidth]{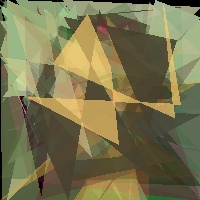} }
    \caption{Samples from polygon tests (10,000 generations), with 5, 25, and 50 polygons.}
    \label{figure:polygonspics}
\end{figure}

\subsection{Circles}
\begin{figure}[t]
    \centering
    \includegraphics[width=\textwidth]{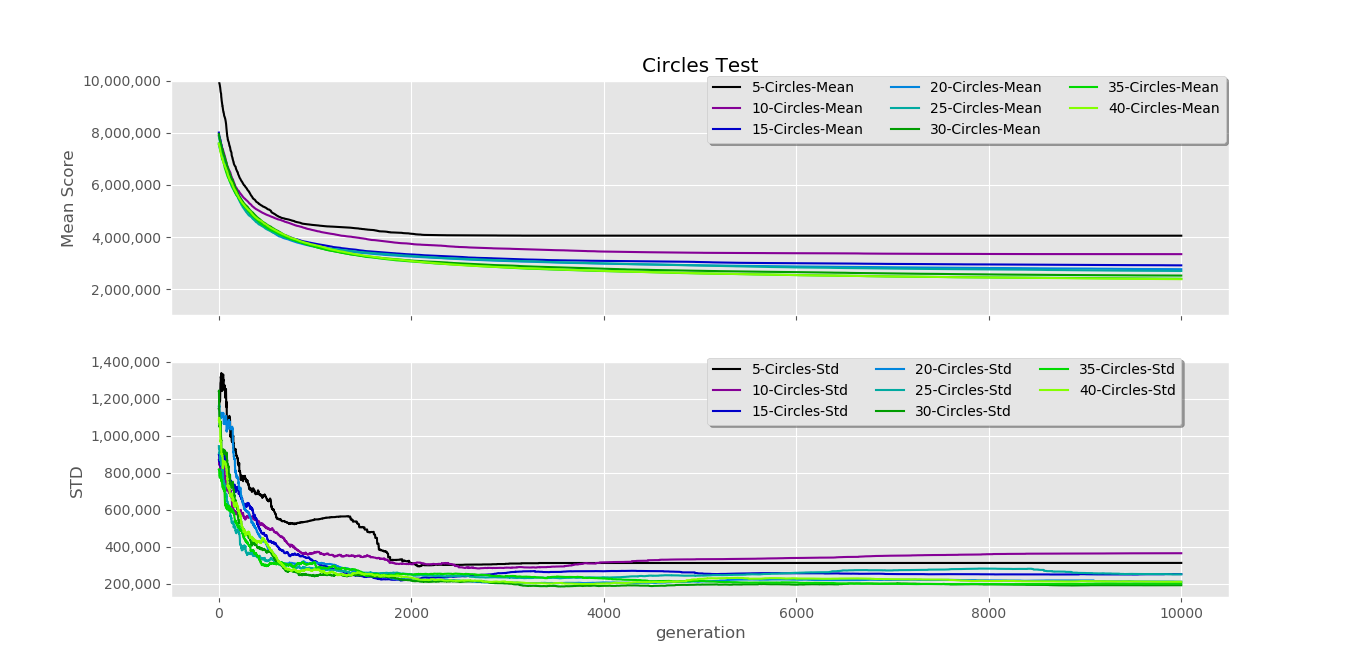} \caption[Result from the circles tests]{Result from the circles tests from 5 to 40 circles. Top graph: Collection of the circles tests. Bottom graph: The STD flattens out already after 2,000 generations. The legend shows number of circles.}
    \label{figure:circtest}
\end{figure}

Using genomes consisting only of circles, testing showed that, similarly to polygons, the fitness score increases steadily when adding more circles to the genome (see Figure \ref{figure:circtest}). An all circles genome gets an average approximation of over 90\% at 15 circles. (Whereas polygons were right below that at 15.) The increases in fitness score when increasing genome complexity are more volatile with circles than polygons. We only tested up to 40 circles per genome and do not know if average fitness score decreases between 40 and 50 circles as with polygons. Using 40 circles was however only slightly better than using 35 (92.20\% over 92.15\% (0.05 percent point increase)), whereas using 30 circles was a great deal better than using 25 (91.75\% over 91.17\% (0.58 percent point increase)). There are no clear winners, but genomes consisting of 15, 20 and 30 circles stand out among the rest.

\begin{figure}[t]
\centering
    {\includegraphics[width=0.25\textwidth]{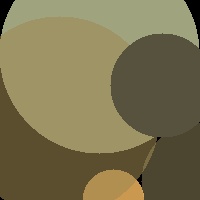}}
    {\includegraphics[width=0.25\textwidth]{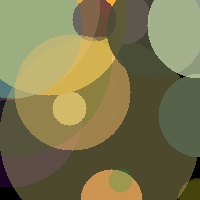}}
    {\includegraphics[width=0.25\textwidth]{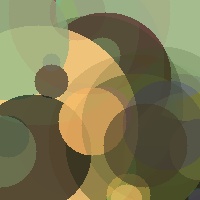} }
    \caption{Samples from the circles tests (10,000 generations), with 5, 20, and 40 circles.}
    \label{figure:circlespics}
\end{figure}

\begin{figure}[t]
\centering
    {\includegraphics[width=0.25\textwidth]{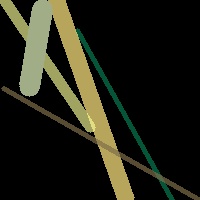}}
    {\includegraphics[width=0.25\textwidth]{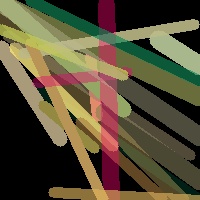}}
    {\includegraphics[width=0.25\textwidth]{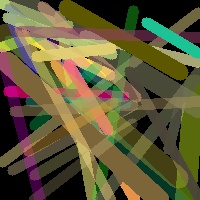}}
    \caption{Samples from the lines tests (10,000 generations), with 5, 20 and 40 lines}
    \label{figure:linespics}
\end{figure}

\begin{figure}[h!]
\centering
    {\includegraphics[width=0.25\textwidth]{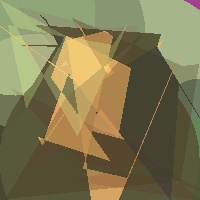}}
    {\includegraphics[width=0.25\textwidth]{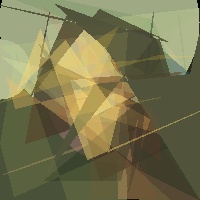}}
    {\includegraphics[width=0.25\textwidth]{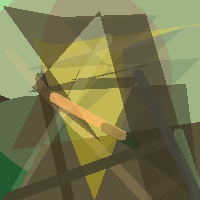}}
    \caption{Samples from the gene combination tests (10,000 generations) with [10 Circles 10 Poly], [5 Circles 15 Polygons], [5 Circles 5 Polygons 10 Lines]}
    \label{figure:polypics}
\end{figure}

\subsection{Lines} \label{section:lines}

Testing with genomes consisting only of lines was shown to be an ineffective option. Starting at an average approximation of 65.50\% using 5 lines in the genome, ending at an average approximation of 83.86\% using 40 lines per genome. While using lines is worse at getting desired fitness results, it may still provide a certain aesthetic and artistic value for human eyes. And may prove useful in approximating certain pictures containing elements with straight lines.

\subsection{Combinations of polygons, circles, and lines}
To determine if a combination of different genes had any advantage over single gene types, we tested 6 different compositions with 20 genes in total. The tests showed that the combination of circles and polygons had the best results. The best composition of genes was a 1:1 ratio of polygons and circles, and the second best a 3:1 ratio of polygons to circles. Based on the experiments, lines make very little impact on the fitness results over the course of 10,000 generations, but they may provide functionality for certain images and artistic purposes.

\subsection{Mutation Probability}

To find the best mutation probability we chose to only do tests on the smallest population size of 1, due to computational time. The result showed that a mutation probability higher than 50\% overall gave poor results over 10,000 generations.

This prompted us to limit the future testing to 10\% and 30\%. A mutation probability of 10\% entails that every single gene has a 10:1 chance of mutating. One genome consisting of 40 genes will \textit{on average} have 4 of its genes mutated every generation. If the probability is 30\%, then the average will be 12 genes per generation.

\section{Other Considerations}

After thorough testing, we have found a collection of \textit{suitable values} for every parameter to produce visually pleasing results. Some of the values disproportionately increase the computation time, and are therefore not beneficial even if they produce a better fitness. Some parameter values that have been omitted might have potential to give aesthetically pleasing or interesting results. Interestingly, some of the parameters we tested did not have the impact we initially thought they would, such as \textit{Lines} and \textit{Crossover}, and have been omitted from the collection. Polygons and circles were both relatively good at around 25 genes, and are therefore both represented at 25 in order to maintain a 1:1 ratio, which was the best distribution of genes in the tests. 

\section{Conclusions}

Our work shows that it is possible to generate artistic images through evolutionary algorithms, and could be used for artistic purposes. We have certainly been amazed and puzzled by the images our algorithm has produced. Some examples are shown below in the Appendix.

%
%
%
\bibliographystyle{splncs04}
\bibliography{bib}
\newpage
\section{Appendix}
\begin{figure}[h]
    \centering
    \includegraphics[width=0.25\textwidth]{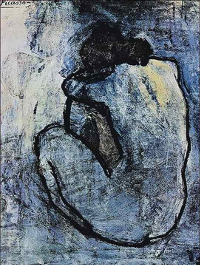}
    \includegraphics[width=0.25\textwidth]{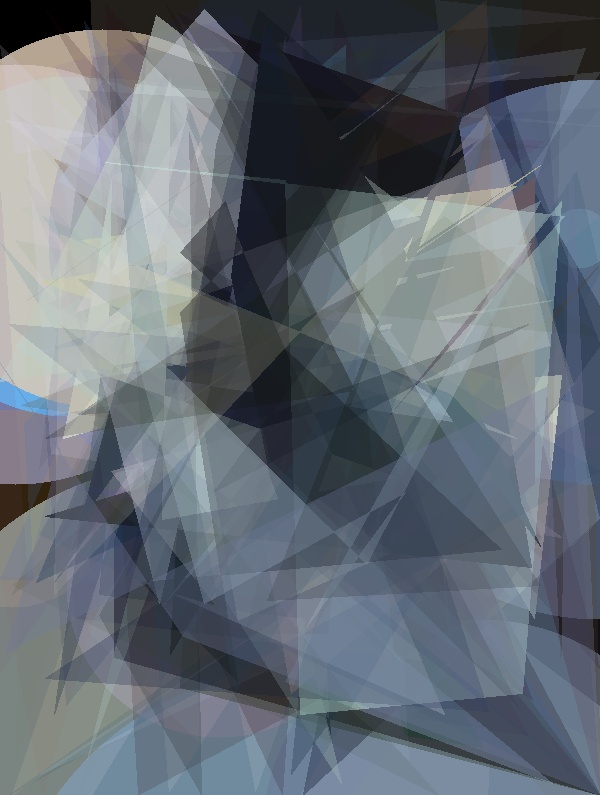}
    \caption{Blue Nude by Pablo Picasso}
\end{figure}

\begin{figure}[h]
    \centering
    \includegraphics[width=0.15\textwidth]{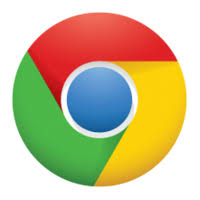}
    \includegraphics[width=0.15\textwidth]{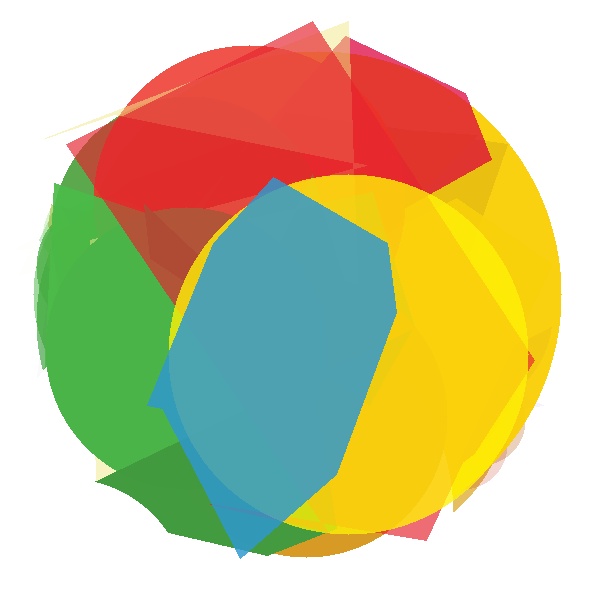}
    \caption{Google Chrome logo}
\end{figure}

\begin{figure}[h]
    \centering
    \includegraphics[width=0.35\textwidth]{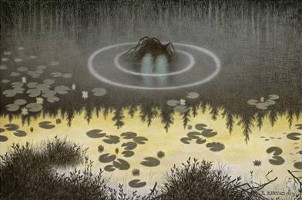}
    \includegraphics[width=0.35\textwidth]{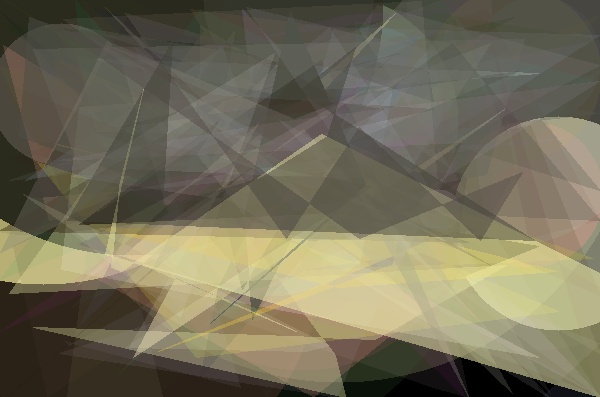}
    \caption{Nokken by Theodor Kittelsen}
\end{figure}

\end{document}